%% file: ms.tex
\documentclass[letterpaper, 10 pt, conference]{ieeeconf}  

\IEEEoverridecommandlockouts                              

\usepackage{graphics} 
\usepackage{epsfig} 
\usepackage{mathptmx} 
\usepackage{times} 
\usepackage{amsmath} 
\usepackage{amssymb}  
\usepackage{subfigure}
\usepackage{amsmath}
\usepackage{amssymb}
\usepackage[space]{cite}
\usepackage{graphicx}
\usepackage{footmisc}

\title{\LARGE \bf
OneShot Global Localization: Instant LiDAR-Visual Pose Estimation
}
\author{
    \IEEEauthorblockN{Sebastian Ratz\IEEEauthorrefmark{1}, Marcin Dymczyk\IEEEauthorrefmark{2}, Roland Siegwart\IEEEauthorrefmark{2}, and Renaud Dub\'{e}\IEEEauthorrefmark{1}}
    \IEEEauthorblockA{\IEEEauthorrefmark{1}Autonomous Systems Lab, ETH Zurich, Switzerland
    \\\{1, 4\}@abc.com}
    \IEEEauthorblockA{\IEEEauthorrefmark{2}Sevensense Robotics AG
    \\\{2, 3\}@sevensense.ch}
}
\author{Sebastian Ratz$^{1,2}$, Marcin Dymczyk$^{1}$, Roland Siegwart$^{2}$, and Renaud Dub\'{e}$^{1}$
\thanks{$^{1}$Sebastian Ratz, Marcin Dymczyk, and Renaud Dub\'{e} are with Sevensense Robotics AG, Zurich, Switzerland (firstname.lastname$@$sevensense.ch).}
\thanks{$^{2}$ Sebastian Ratz and Roland Siegwart are with the Autonomous Systems Lab, ETH Zurich, Switzerland.}
}

\begin{document}

\input{chapters/title_page}
\maketitle
\thispagestyle{empty}
\pagestyle{empty}

\input{chapters/abstract}
\input{chapters/introduction}
\input{chapters/related_work}

\input{chapters/method_lidar}
\input{chapters/method_vision}
\input{chapters/experiments}
\input{chapters/conclusion}


 \addtolength{\textheight}{-2cm}

\bibliographystyle{IEEEtran}
\bibliography{bibliography/references}
\end{document}

%% file: chapters/title_page.tex
\par\smallskip\noindent
\begin{minipage}{\textwidth}
\begin{center}
\vspace{30pt}

This paper has been accepted for publication at: 

\textit{IEEE International Conference on Robotics and Automation (ICRA), 2020.}

\vspace{30pt}

Please cite our work as:

\vspace{12pt}
S. Ratz, M. Dymczyk, R. Siegwart, R. Dub{\'e}. ``OneShot Global Localization: Instand LiDAR-Visual Pose Estimation''. \textit{IEEE International Conference on Robotics and Automation (ICRA).}
\vspace{30pt}
\end{center}

bibtex:
\begin{verbatim}
@inproceedings{oneshot2020ratz,
author = {Sebastian Ratz and Marcin Dymczyk and Roland Siegwart and Renaud Dub{\'e}},
title = {One{S}hot Global Localization: Instant LiDAR-Visual Pose Estimation},
booktitle = {IEEE International Conference on Robotics and Automation (ICRA)},
year = {2020}
}
\end{verbatim}
\end{minipage}
\par\smallskip

\clearpage

%% file: chapters/abstract.tex
\begin{abstract}
Globally localizing in a given map is a crucial ability for robots to perform a wide range of autonomous navigation tasks. This paper presents \textit{OneShot}~--~a global localization algorithm that uses only a single 3D LiDAR scan at a time, while outperforming approaches based on integrating a sequence of point clouds. Our approach, which does not require the robot to move, relies on learning-based descriptors of point cloud segments and computes the full 6 degree-of-freedom pose in a map. The segments are extracted from the current LiDAR scan and are matched against a database using the computed descriptors. Candidate matches are then verified with a geometric consistency test. We additionally present a strategy to further improve the performance of the segment descriptors by augmenting them with visual information provided by a camera. For this purpose, a custom-tailored neural network architecture is proposed. We demonstrate that our LiDAR-only approach outperforms a state-of-the-art baseline on a sequence of the KITTI dataset and also evaluate its performance on the challenging NCLT dataset. Finally, we show that fusing in visual information boosts segment retrieval rates by up to 26\% compared to LiDAR-only description.
\end{abstract}

%% file: chapters/introduction.tex
\section{Introduction}
\label{sec:introduction}

The ability of a robot to determine its 6 degree-of-freedom (DOF) pose within an environment is a critical element in Simultaneous Localization and Mapping (SLAM). Modern SLAM algorithms often depend on reliable and accurate place recognition algorithms in order to perform global localization and loop closure detection \cite{Cadena2016PastPresentFuture}. These enable robots to find their global pose in a map and can be used to correct for drift of the local odometry estimators. Finally, SLAM is key to performing a wide range of tasks including navigation, inspection, and maintenance. 

Global localization using 3D Light Detection and Ranging (LiDAR) sensors has attracted significant research interest due the sensors' independence of the appearance of the environment. This brings an advantage over cameras in ill-lighted conditions or in situations where illumination undergoes strong changes over time \cite{VisionSurveyLowry}. Global localization using LiDAR sensors is however a challenging task -- due to the sparse nature of the data and computational complexities -- and is still regarded as an open problem. Among recent techniques addressing this problem, our work builds upon point cloud segment extraction and matching strategies~\cite{oxfordSegments2,segmatch_2017, segmap2018}.

\begin{figure}
   \centering
   \includegraphics[width=1.0\columnwidth]{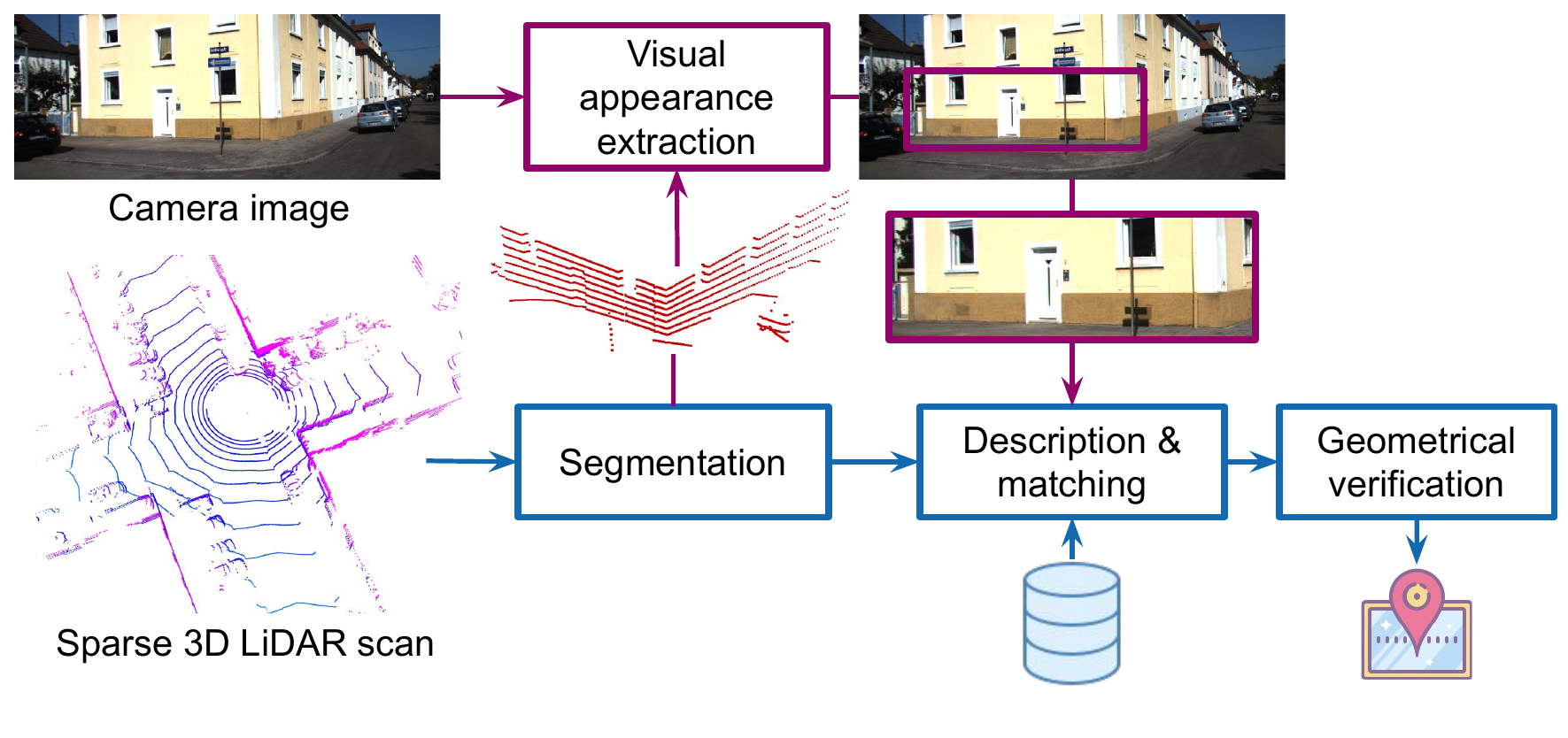}
   \caption{Our global localization pipeline which uses a single sparse 3D LiDAR scan at a time is outlined in blue. Segments are extracted from a LiDAR scan, described using a neural network and matched against a database. The correspondences are then verified and used to estimate the global 6 DOF pose of the robot. Optionally, depicted in purple, we propose a method to augment point cloud segments with the visual information encoded in overlapping camera images, in order to enhance the segment descriptor performance.}
   \label{fig:schema}
   \vspace{-5mm}
\end{figure}
Our first contribution is to extend on these existing methods by introducing \textit{OneShot}: a global localization approach which does not require the robot to move for localizing itself in a map.\footnote{A video providing an algorithmic overview and impressions of our experiments in challenging real-world scenarios is available at: \url{https://youtu.be/EHdzBu7pYrc}} This mitigates the potential risks associated with navigating under high uncertainty of the global robot pose. In order to obtain dense point clouds without a moving robot, it has been proposed to actuate LiDAR sensors mechanically~\cite{SpringMount, Cop2018DelightAE}. Contrastingly, our approach is free of such sensor actuation mechanisms, which eliminates the associated complexity and increases the robustness of the localization system. Furthermore, our proposed algorithm brings the advantage of being independent from inherently noisy pose estimates required for point cloud integration, which can have a negative impact on the quality of the segment descriptors. 

While the frame-by-frame principle is very common in vision systems \cite{Cadena2016PastPresentFuture} it is more challenging to handle in the case of LiDAR data, due to their sparse nature. We address this challenge with a novel global localization approach and are able to demonstrate superior performance compared to algorithms which rely on point cloud integration prior to segmentation. Finally, we show in several experiments that \textit{OneShot} works with sparse 16-beam LiDAR data and executes global localization in real-time using a consumer-grade GPU.

As many robotic platforms use a combination of LiDAR and vision sensors, we propose a strategy to combine the strengths of LiDAR and Vision sensors in order to further increase the retrieval performance of the descriptors, as depicted in Fig.~\ref{fig:schema}. This is achieved by augmenting the structural information of point cloud segments with the visual appearance encoded in images, thus bringing together two complementary sensor modalities. Our experiments show that fusing LiDAR and visual information significantly improves the distinctiveness of our segment descriptors.

To summarize, this paper provides the following contributions:
\begin{itemize}
  \item \textit{OneShot}: A novel 3D LiDAR-based global localization approach which uses a single scan for estimating a robot's 6-DOF pose.
  \item A LiDAR-Vision segment descriptor based on a custom-tailored neural network, significantly improving over the distinctiveness of LiDAR-only descriptors.
  \item An extensive evaluation of the presented algorithms on publicly available datasets demonstrating increased performance over state-of-the-art baselines.

\end{itemize}


%% file: chapters/related_work.tex
\section{Related Work}
\label{sec:related_work}
Global localization in LiDAR data is a challenging problem, which was addressed by a considerable amount of research works. We found that most methods can be classified into three categories: keypoint-based approaches, global descriptor methods and strategies based on intermediate-sized representations such as segments, objects or geometric primitives such as planes or edges.

Keypoint based approaches such as~\cite{fpfh, SHOT, NARF} are based on extracting and matching local point cloud descriptors. A popular keypoint descriptor using Fast Point Feature Histograms (FPFH) \cite{fpfh} is based on encoding the geometrical relations of the $k$-neighborhood of a point in histograms. A work by Steder et al.~\cite{NARF} introduced Normally Aligned Radial Features (NARF) which are computed directly on range images. Signatures of Histograms for Local Surface Description (SHOT) \cite{SHOT} quantify local surface normal information for improved repeatability.
All these methods, however, rely on dense point data and may not work well with sparse point clouds as produced by sensors like the Velodyne VLP-16.

Working at the level of sparse point clouds for global localization has also been proposed in the vision community by Cieslewski et al.~\cite{visionSparseCieslewski}. After triangulating visual features to sparse point clouds, the authors propose to compute 3D local point cloud features based on point density comparisons of neighboring regions. A more recently presented work \cite{learnedKP2018} introduces a learning based approach using a Siamese architecture and is able to cope with sparse 3D LiDAR data. The authors report competitive results in terms of matching accuracy and computation times compared to SHOT and FPFH, for the task of aligning consecutive LiDAR scans. Contrastingly, our approach can be applied for associating any two given scans, i.e. solving the full prior-free global localization problem.

Since keypoint-based approaches are limited by the lack of distinctiveness of local structures in LiDAR data, global point cloud descriptors represent an alternative and have enjoyed vivid interest by the research community in the recent years. Both, hand-crafted and learning based approaches have been presented. Among the hand-crafted descriptors, Röhling et al.~\cite{heightHistogram2015} propose a global descriptor based on histograms encoding the point height distribution of a point cloud. He et al.~\cite{M2DP} presented \textit{M2DP}, a global point cloud descriptor based on projecting the input cloud on multiple 2D planes and computing a density signature from the projections. 

Active research in the field of learning based global point cloud descriptors has produced various promising approaches in the recent years \cite{Cop2018DelightAE,PointNetVLAD2018,LocNet2017,PointCNN2018,PointNetPP2017}. 
Cop et al.~\cite{Cop2018DelightAE} presented \textit{DELIGHT}, a global point cloud descriptor which leverages the intensity information which is available for multiple recent LiDAR sensor models, additionally to the point cloud structure. In contrast to our approach, their system requires dense point clouds to function and actuates a LiDAR sensor for this purpose. Qi et al. \cite{PointNet2016,PointNetPP2017} presented architectures computing descriptors directly from unorganized point clouds, called \textit{PointNet} and \textit{PointNet++}. The latter extended the capabilities of its predecessor to capture fine-grained structures in point clouds. \textit{PointNetVLAD} is a neural network architecture based on \textit{PointNet} \cite{PointNet2016} and the network architecture \textit{NetVLAD} \cite{NetVLAD2015}, using the latter to aggregate point cloud features in a VLAD-like descriptor. 
In comparison to their hand-crafted counterparts, these learned approaches are often shown to be more robust and informative. However, global descriptors rely on the assumption that places are revisited from similar viewpoints, making them sensible to deviations from the database path. Furthermore, these approaches cast the problem of place recognition as a database retrieval problem only, by finding the closest matching LiDAR scan in a database. In contrast, our algorithm computes the full 6-DOF pose in the given map.

The third group of global localization system relies on descriptors computed from subsets of point clouds which are considerably larger than those considered for describing keypoints. The aim usually is to combine the advantages of both approaches: retaining the descriptiveness of larger point sets while avoiding the issues with invariances of global descriptors~\cite{VisionSurveyLowry}.
Work by Serafin et al.~\cite{fastRobustKp2016} proposes to extract plane and edge structures based on principal component analysis (PCA) of local point cloud regions. In our previous work, we proposed to extract, describe and match point cloud segments, without expecting predefined geometries such as cylinders, planes or edges \cite{segmatch_2017}. The point cloud description technique was later extended to \textit{SegMap}: a data-driven approach that uses a 3D convolutional neural network for description~\cite{segmap2018, dube2019segmap}. Other variants of this segment-based approach were presented by Tinchev et al.~\cite{oxfordSegments1}, \cite{oxfordSegments2} which similarly depend on robot movement and point cloud integration.

Our work extends \textit{SegMap} and improves both its applicability and performance. Current state-of-the-art segment based approaches accumulate point clouds into a local map and try to localize it in a global map. They thus require the robot to blindly explore its vicinity to localize or the sensor to be actuated. Our algorithm, in contrast, is able to localize on a frame-by-frame basis. Localization can thus be done sooner after startup, minimizing potentially unsafe dead-reckoning phases.

%% file: chapters/method_lidar.tex
\section{One-Shot Lidar Localization}
\label{sec:method_lidar}
This section describes the the proposed global localization algorithm outlined in Figure~\ref{fig:schema}. We propose to extract segments from a single LiDAR scan and describe them using a neural network trained specifically for this task. A matching and geometrical verification step results in the output of a full 6 DOF pose estimate in the map. Each of the modules is detailed below and together they allow a robotic system to perform effective one-shot global localization using a LiDAR sensor, without performing the costly data integration.

\subsection{Segmentation}
\begin{figure}
   \centering
é   \includegraphics[width=1.0\columnwidth]{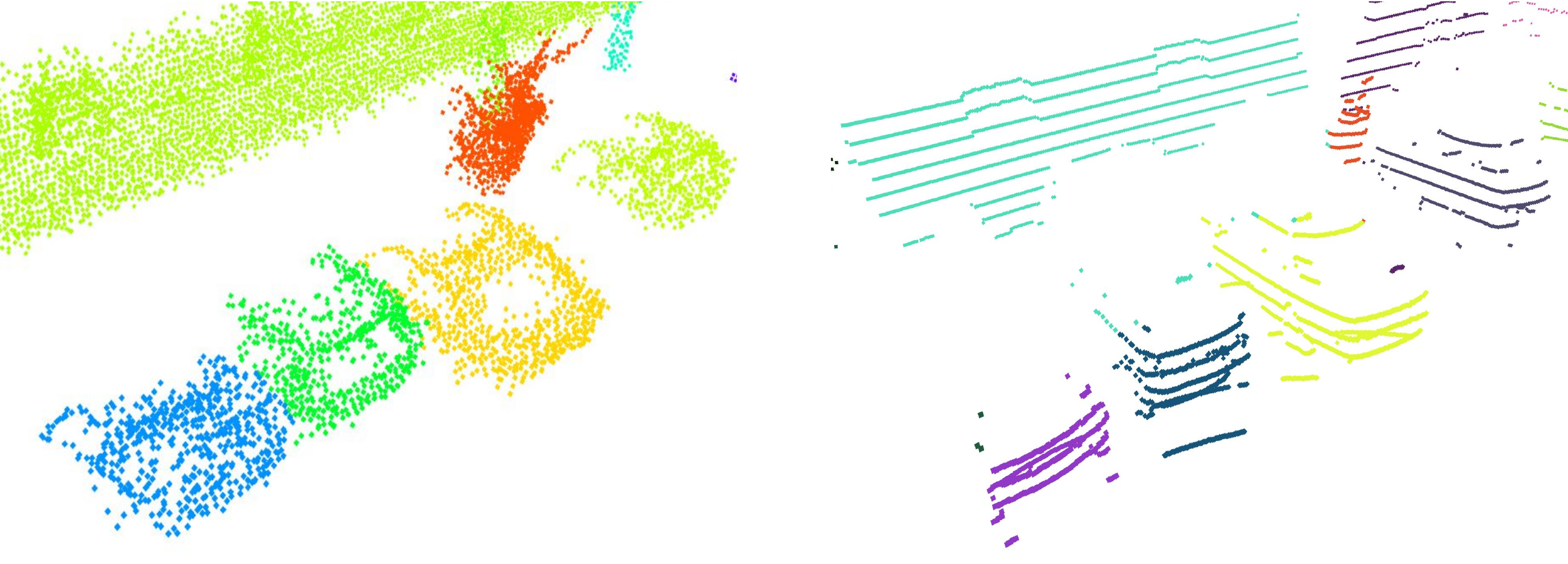}
   \caption{Left: State-of-the-art segment based global localization approaches extract dense segments from accumulated point clouds~\cite{segmatch_2017, oxfordSegments2}. Every segment is colored differently. Right: In this work we rely on segments extracted from a single sparse 3D LiDAR scan. Apart from the lower point density, the sparse segments tend to represent smaller parts of the underlying structure, as can for example be seen with the red tree trunk. However, single-scan segments are free of the noise introduced by point cloud accumulation and thus have a sharper geometry.}
   \label{fig:sparse_dense}
   \vspace{-4mm}
\end{figure}

One of the main differences to previous segment based approaches \cite{segmatch_2017, segmap2018, oxfordSegments1, oxfordSegments2} is that our system works with a single LiDAR scan at a time. When additionally dealing with sparse LiDAR data, segmentation becomes a challenging task. A comparison between segments extracted from an accumulated dense point cloud with those from a single sparse LiDAR scan is shown in Figure~\ref{fig:sparse_dense}. Sparse LiDAR data are provided by scanners such as those from Velodyne\footnote{\url{https://velodynelidar.com/vlp-16.html}} or Ouster\footnote{\url{https://ouster.com/products/os1-lidar-sensor}}. These may have as little as 16 scan lines. Euclidean distance or curvature based segmentation approaches which are typically used often fail when confronted with such sparse data, since distances between points of two different scan lines are usually large and point normals may thus be inaccurate. The specific reading pattern of the sensor would significantly affect the segmentation if not addressed explicitly. For this reason, we leverage the segmentation approach presented by Bogoslavskyi et al.~\cite{DepthClustering}. This efficient algorithm works directly on the range image and employs a simple geometrical strategy to separate structures in the point cloud which exhibit strong discontinuity in depth.

\subsection{Segment Description}
The objective of the segment description module is to perform dimensionality reduction on the incoming point cloud segments, compressing them to a vector of floating point numbers. In our experiments, 64 proved to be a good size for the embedding. 

The description is performed by an artificial neural network based on 3D convolutions, max pooling and dense layers. This architecture was first presented by  Dub\'{e} et al.~\cite{segmap2018} and resembles closely what is used in the presented algorithm. Since our approach processes point clouds from a single scan in contrast to data accumulated from multiple scans, the network was retrained on the sparse data. We also took advantage of recent advances in metric learning and implemented the triplet loss~\cite{Facenet2015} for training the network.

\subsection{Neural Network Training}
\label{sec:training}
In the following, the method employed for training the artificial neural network is explained. First, the training data generation is described, followed by an explanation of the used loss function and training technique. 

\subsubsection{Training Data Generation}
\label{sec:data_generation}
\begin{figure}
   \centering
   \includegraphics[width=1.0\columnwidth]{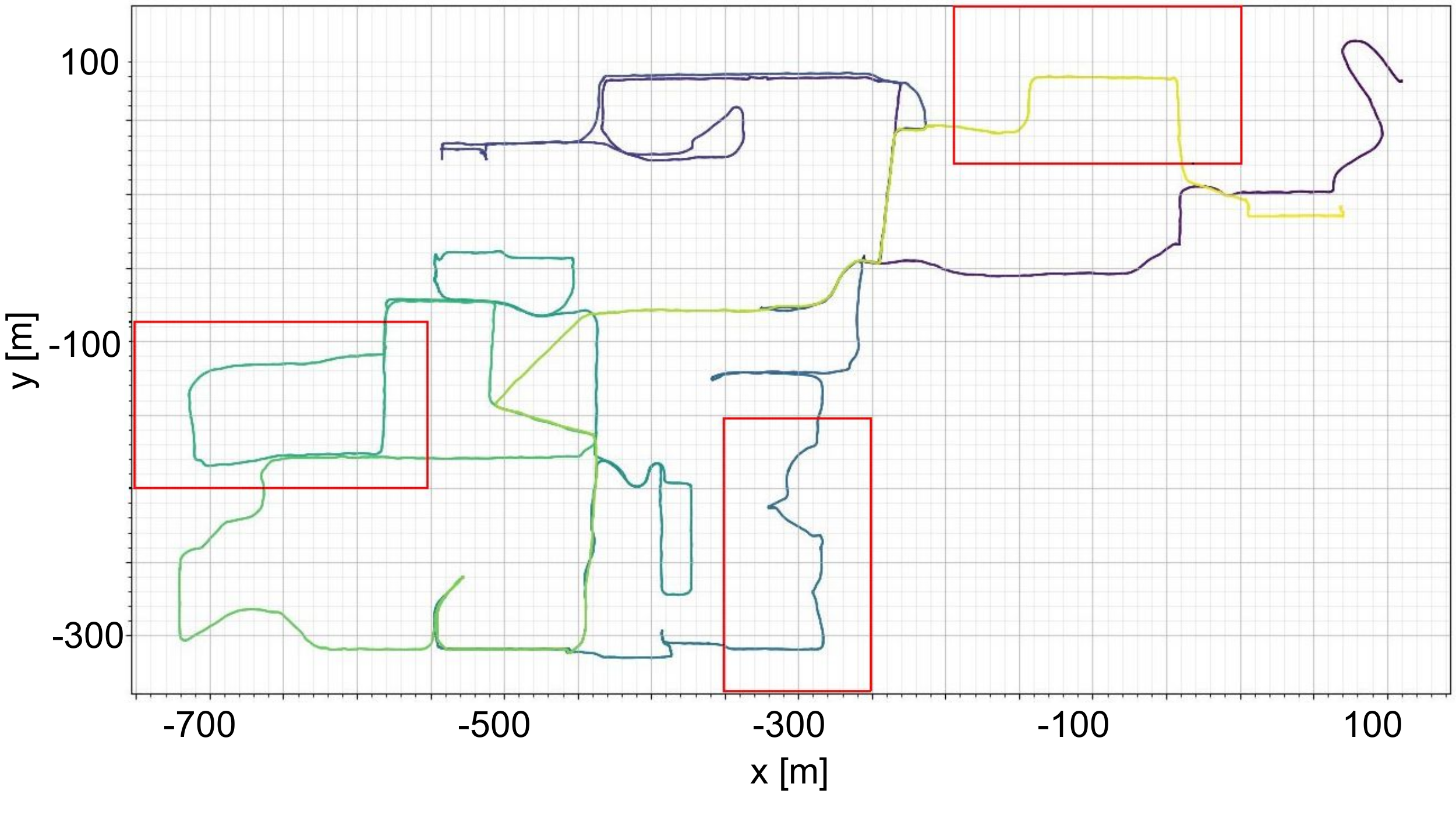}
   \caption{The image shows the trajectory recorded on 2012-11-17 of the NCLT dataset. The trajectory changes color with time, starting at purple and going over to blue, green and yellow.  Data recorded within the three red boxes are exclusively used for testing, while all other data are used for training the neural network. The test set was selected such that it contains both wide open spaces and narrow, urban-like areas.}
   \label{fig:data_separation}
   \vspace{-4mm}
\end{figure}

The training data is generated from the public datasets KITTI~\cite{Kitti} and NCLT~\cite{NCLT}. A pair of segments is marked as a ground truth correspondence, if the intersection of their convex hulls is large in comparison to their union \cite{segmap2018}. For the KITTI odometry dataset~\cite{Kitti}, sequence 05 and 06 are used for training and sequence 00 for testing. In the case of the NCLT dataset, the largely overlapping recordings from the dates 2012-11-17 and 2012-03-25 were used. The separation of training and test data for the NCLT dataset is outlined in Figure~\ref{fig:data_separation}.

\subsubsection{Loss Function}
The triplet loss \cite{Facenet2015} is a method for metric learning which has gained considerable popularity for the application of visual place recognition \cite{Cieslweski2017, Arandjelovic16Netvlad, Loquercio2017} and is employed in this work for the task of finding embeddings for 3D point cloud segments. A crucial element in triplet training is an appropriate strategy for hard-negative mining, which has been shown to improve performance and convergence rates \cite{Facenet2015, hierarchicalTriplet2018, SamplingMatters2017}. We chose a strategy called \textit{batch hard}~\cite{batchHard2017} for its balance between finding challenging hard samples without becoming too sensitive to mislabelled samples in the training set.
 
 \subsection{Segment Matching}
Our localization map consists of a database containing the segments obtained during the mapping phase, which are stored together with their centroid positions. Segment descriptors are maintained in a k-d tree structure to allow for efficient nearest neighbor search. For every segment extracted from a query scan, its $k$ nearest neighbors in descriptor space are retrieved from the map and passed into the next section of the pipeline.
 
 \subsection{Geometric Consistency Test}
 After segment description and matching, the candidates are passed through the geometric consistency test. The implementation follows closely the graph formulation approach presented by Dub\'{e} et al.~\cite{incremental2018}. The algorithm finds the maximum clique in a so called \textit{consistency graph}. The result corresponds to the maximum set of segment matches which have a consistent configuration in the query scan and the global map. Finally, solving a least squares problem minimizing the distances between consistent map and database segment centroids allows us to compute the global 6 DOF pose~\cite{incremental2018}.

%% file: chapters/method_vision.tex
\section{Method: Augmenting with Vision}
\label{sec:method_vision}
This section describes our strategy to enhance point cloud segment descriptors with visual information. We pursue the objective of augmenting the structural information contained in LiDAR data with the visual appearance of the environment provided by camera sensors.

\subsection{Image Patch Extraction}
Our approach requires that a camera has significant overlap with the field of view of the LiDAR sensor. Once an incoming point cloud has been segmented, the resulting segments are projected onto the camera image. Following this, the bounding box of the projection is computed and the corresponding part of the image is extracted. The resulting image patch represents the visual appearance of the corresponding segment and serves to complement the structural information encoded in the LiDAR data.

\subsection{LiDAR-Vision Fusion}
For the description of the image patch we use the neural network NetVLAD~\cite{NetVLAD2015}. It is a neural network architecture which became a \textit{de facto} standard for visual global localization and produces a global descriptor of an image. An incoming image is passed through a deep convolutional network based on VGG-16 \cite{vgg16} followed by a NetVLAD layer. The latter aggregates the features coming from the convolutional network into learned clusters and produces a single $4096\times1$ descriptor of the image. An optional PCA-based dimensionality reduction may be applied, however we do not make use of this last feature in our work. The publicly available weights of an implementation by Cieslewski et al.~\cite{Cieslewski2017DataEfficientDV} were used and kept fixed during the training of the multi-modal descriptor network.

\begin{figure}
   \centering
   \includegraphics[width=0.8\columnwidth]{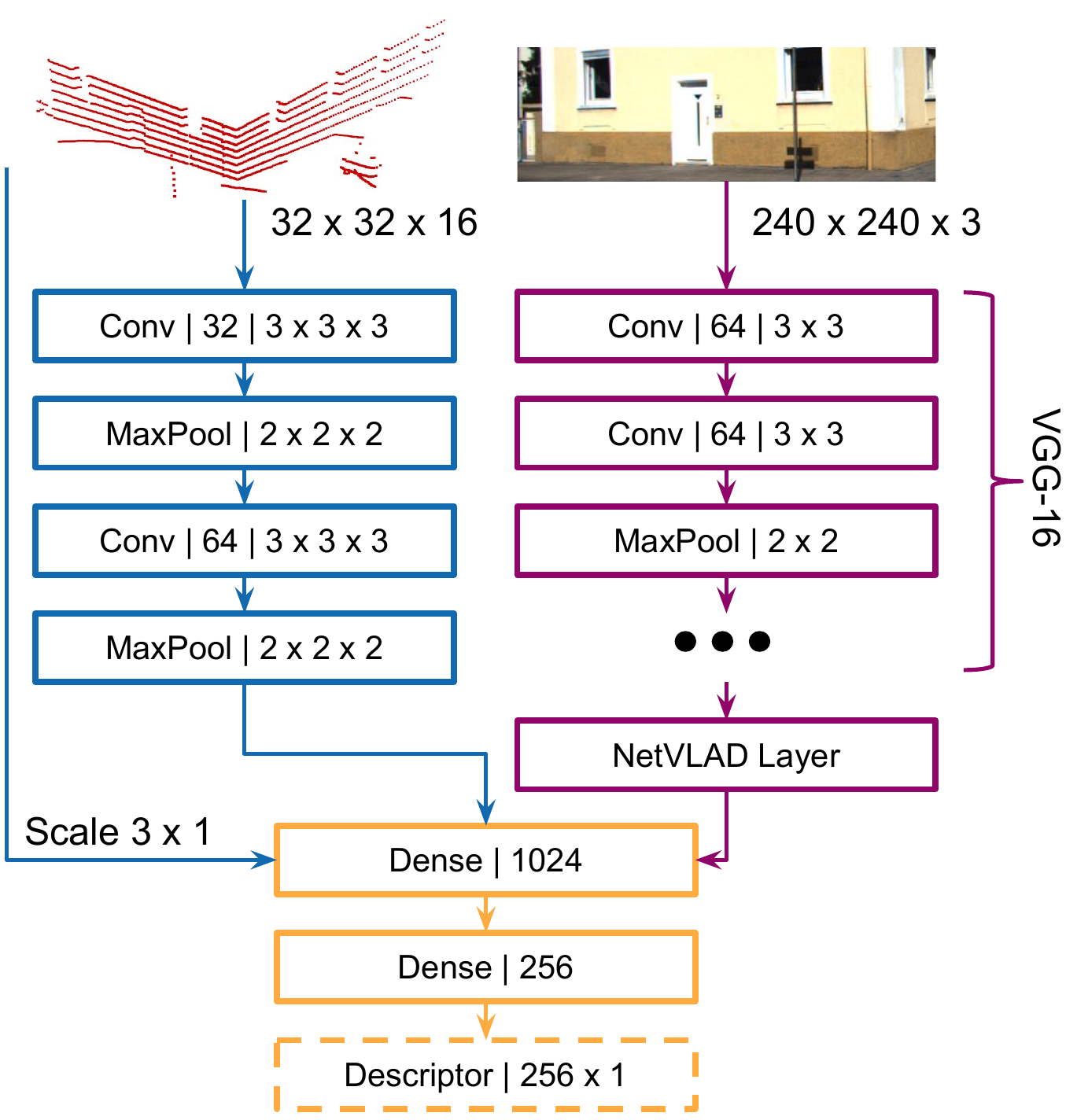}   \caption{The neural network proposed to compute the multi-modal descriptor. In blue, the layers for computing the intermediate LiDAR embedding are depicted. The purple layers perform the image patch description based on NetVLAD~\cite{NetVLAD2015}. Finally, the embeddings are fused using dense layers.}
   \label{fig:nn_diagram_fusion}.
   \vspace{-4mm}
\end{figure}

A detailed diagram of the neural network structure developed for computing a multi-modal descriptor of the segment and the image is depicted in Figure~\ref{fig:nn_diagram_fusion}. In our use of NetVLAD, the image patch is first re-scaled to a size of 140x140 and then fed into the network. The resulting descriptor is concatenated with the flattened output of the segment descriptor network. In this case, the segment descriptor network is equal to the convolutional part of the network presented by Dub\'{e} et al.~\cite{segmap2018}. These concatenated intermediate descriptors are then passed through dense layers which finally result in the fused, condensed multi-modal descriptor. Experiments showed that the configuration depicted in Figure~\ref{fig:nn_diagram_fusion} yields a decent trade-off between descriptor performance and compute times.

Training is performed following the same approach as described in Section~\ref{sec:training} with the only difference being the fact that a data sample here consists of a segment point cloud together with a corresponding image patch. The resulting descriptor can be used as a drop-in replacement in the global localization pipeline presented in Section~\ref{sec:method_lidar}.

%% file: chapters/experiments.tex
\section{Experiments}
\label{sec:experiments}

In this section we present the experiments which were executed in order to assess the performance of the proposed solutions. First, the LiDAR-only global localization system is evaluated, followed by an evaluation of the multi-modal LiDAR-Vision descriptor.

\subsection{LiDAR Global Localization}
A quantitative evaluation of the global localization system has been executed on the public datasets KITTI \cite{Kitti}, which represents a urban driving scenario, and NCLT \cite{NCLT}, a long-term and large-scale dataset.

All experiments were executed on an Intel Core i7-8700 CPU and a NVIDIA GeForce RTX 2080 GPU.
\subsubsection{Evaluation on KITTI}
\label{sec:kitti}
Multiple experiments were performed on sequence 00 of the KITTI dataset, which contains a section of 500m length which is visited twice. It is a commonly used benchmark sequence \cite{segmap2018, LocNet2017}. In particular, we used the scans of seconds 340 to 397 as queries and those of seconds 0 to 300 for map building. Note that we did not make use of scans of seconds 300-340 in order to make sure that map and query data are well separated in time. 

In order to demonstrate the ability of our system to work with sparse LiDAR data, we used only 16 of the 64 LiDAR beams provided by KITTI. We chose the channels such that the vertical field of view and resolution is approximating that of a Velodyne VLP-16. The generalization abilities of our segment descriptor network are shown by training it on run 2012-11-17 of the NCLT dataset (see Section~\ref{sec:nclt}).

\begin{figure}
\centering   
        \subfigure {
        \includegraphics[width=0.6\columnwidth]{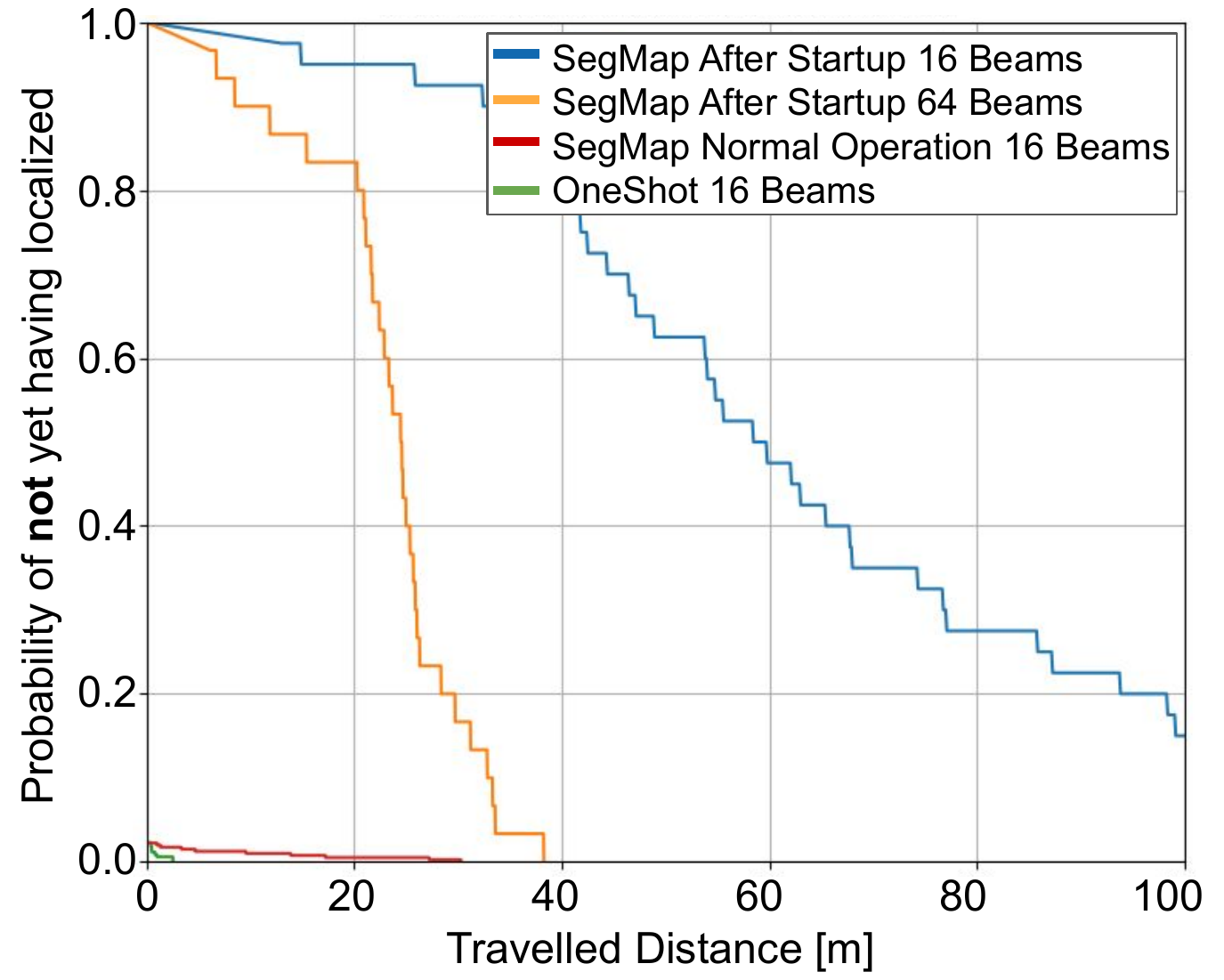}
        }
        \subfigure {
        \includegraphics[width=0.6\columnwidth]{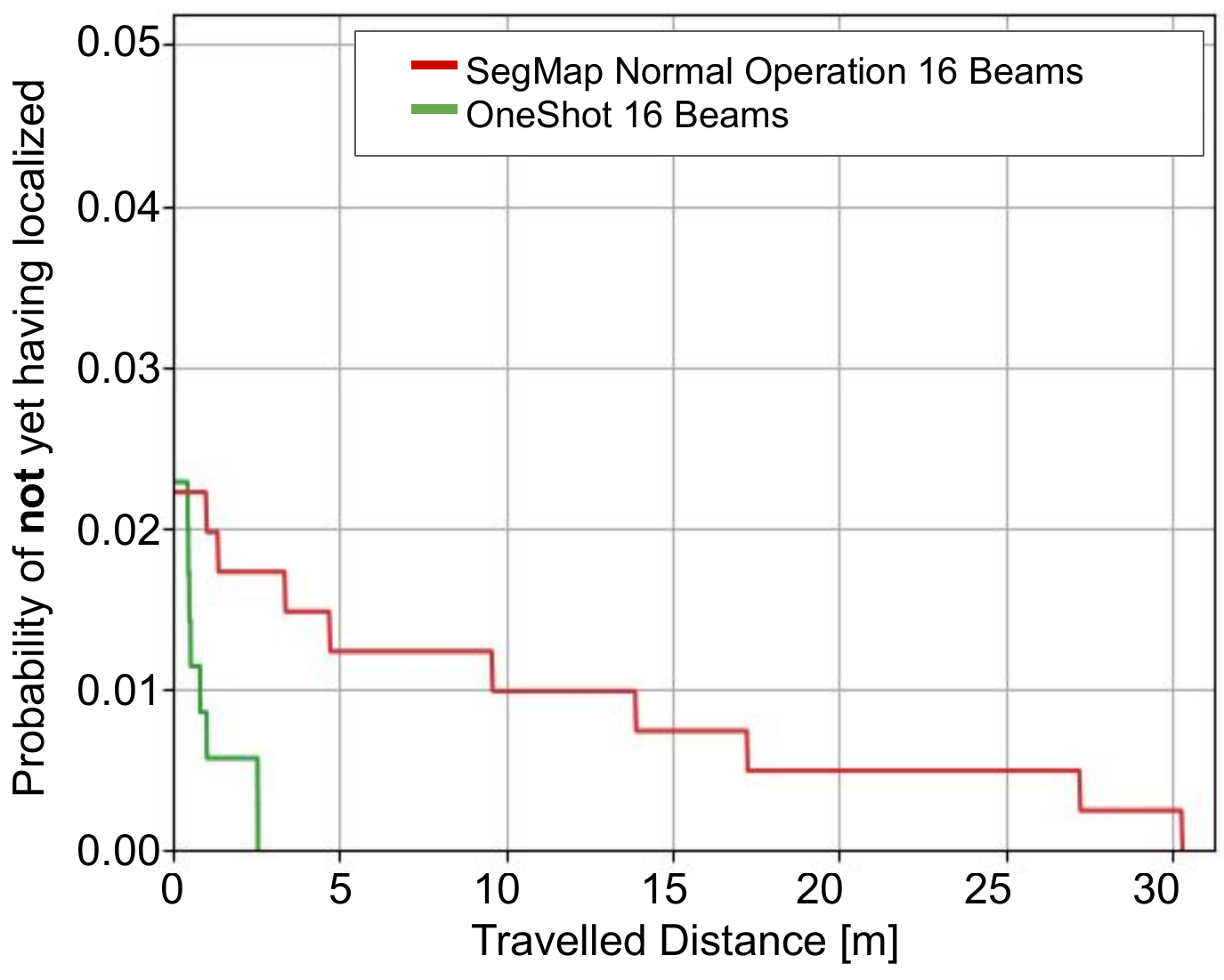}
        }
\caption{The experiment simulates a robot wake-up scenario. We estimate the probability of travelling a given distance without localizing, when starting at an arbitrary frame of the target KITTI sequence 00.  Top: comparison of our approach and the baseline \textit{SegMap}~\cite{segmap2018}.  Bottom: zoomed-in view of the same chart, comparing against \textit{SegMap} during continuous operation.
Shorter distances traveled blindly before localization contribute to increased robustness and safety.}
\label{fig:locdist}
\vspace{-4mm}
\end{figure}
For an unlocalized robot, it is crucial to minimize the dead-reckoning distance required to obtain a global pose, for example after startup-up or when loosing track of the map. Travelling blindly may be unsafe and a time consuming task, hindering the robot's efficiency. To simulate this use-case, we measured the distance travelled after system start-up until localizing for the first time. We repeated this experiment from different starting points along the target sequence, which allows us to compute a probability of travelling a given distance without localizing. We compare to the baseline \textit{SegMap} \cite{segmap2018} which is a state-of-the-art segment based global localization algorithm integrating point clouds in a local map for localization. We ran \textit{SegMap} 40 times, every time starting from a different place in the section. Since \textit{OneShot} relies on single scans only, it was enough to run it through all the scans once in order to compute the same metric. The outcome of the experiment is illustrated in Figure~\ref{fig:locdist}.  Overall, we can report 98\% recall at 100\% precision on the target section of the KITTI dataset, while counting every localization within 2m of the ground truth as a successful localization. This clearly outperforms the baseline \textit{SegMap}, which requires the robot to travel 58m for a localization probability of 50\%. In this experiment, our approach achieved 100\% localization probability after 2.5m while \textit{SegMap} in continuous operation travels up to 30m before localizing.

\begin{figure}
\centering   
        
        \includegraphics[width=0.6\linewidth]{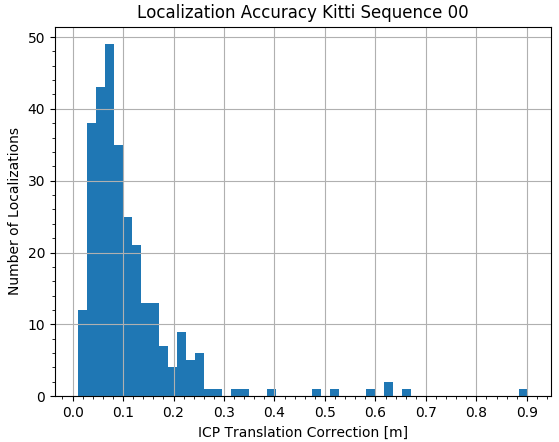}
\caption{Histogram of the localization errors on KITTI sequence 00.
A significant number of segment matches makes the 6-DOF pose estimates relatively stable, with most corrections below 0.3 m.}  
\label{fig:accuracy}
\vspace{-4mm}
\end{figure}

We evaluated the localization accuracy based on the Iterative Closest Point (ICP) algorithm, as was done for our baseline \textit{SegMap}~\cite{dube2019segmap}. For every successful localization of our system, we use the pose estimate as an initial guess for ICP. The computed refinement transformation is then counted as our localization error. Figure~\ref{fig:accuracy} shows a histogram of the performed ICP corrections. The error was computed as 0.11 $\pm$ 0.1 m on average, which compares well to our baseline \textit{SegMap}, for which 0.13 $\pm$ 0.06 m were reported~\cite{dube2019segmap}.

\subsubsection{Evaluation on NCLT}
\label{sec:nclt}
In order to evaluate our approach in a long-term and large-scale setting, we executed experiments on the \textit{The University of Michigan North Campus Long-Term Vision and LIDAR Dataset} (NCLT) \cite{NCLT}. 

We assessed the localization performance on the recording from 2012-03-25 for mapping and that from 2012-08-20 as query. While the campus shows no significant foliage in March, the opposite is true for the recording in August. We chose the same parameters as in the KITTI dataset, except driver specific parameters for the different LiDAR sensor. This time we trained the segment descriptor on KITTI sequence 05 and 06, in order to demonstrate the generalization capabilities of our approach. We report a localization rate of 38\% with a precision of 94\%, demonstrating that our system is also able to localize often on the challenging NCLT dataset. From our observations, the main factors for this are twofold. Firstly, the long-term character of the dataset leads to structural changes between recordings. Especially changing vegetation rendered segments of trees and bushes as unreliable. Secondly, the dataset contains long stretches through open spaces, which pose a great challenge, since little structural information is returned by the LiDAR sensor.

\subsection{Deployment of LiDAR Global Localization}
\begin{figure}
\centering   
        \includegraphics[width=1.0\linewidth]{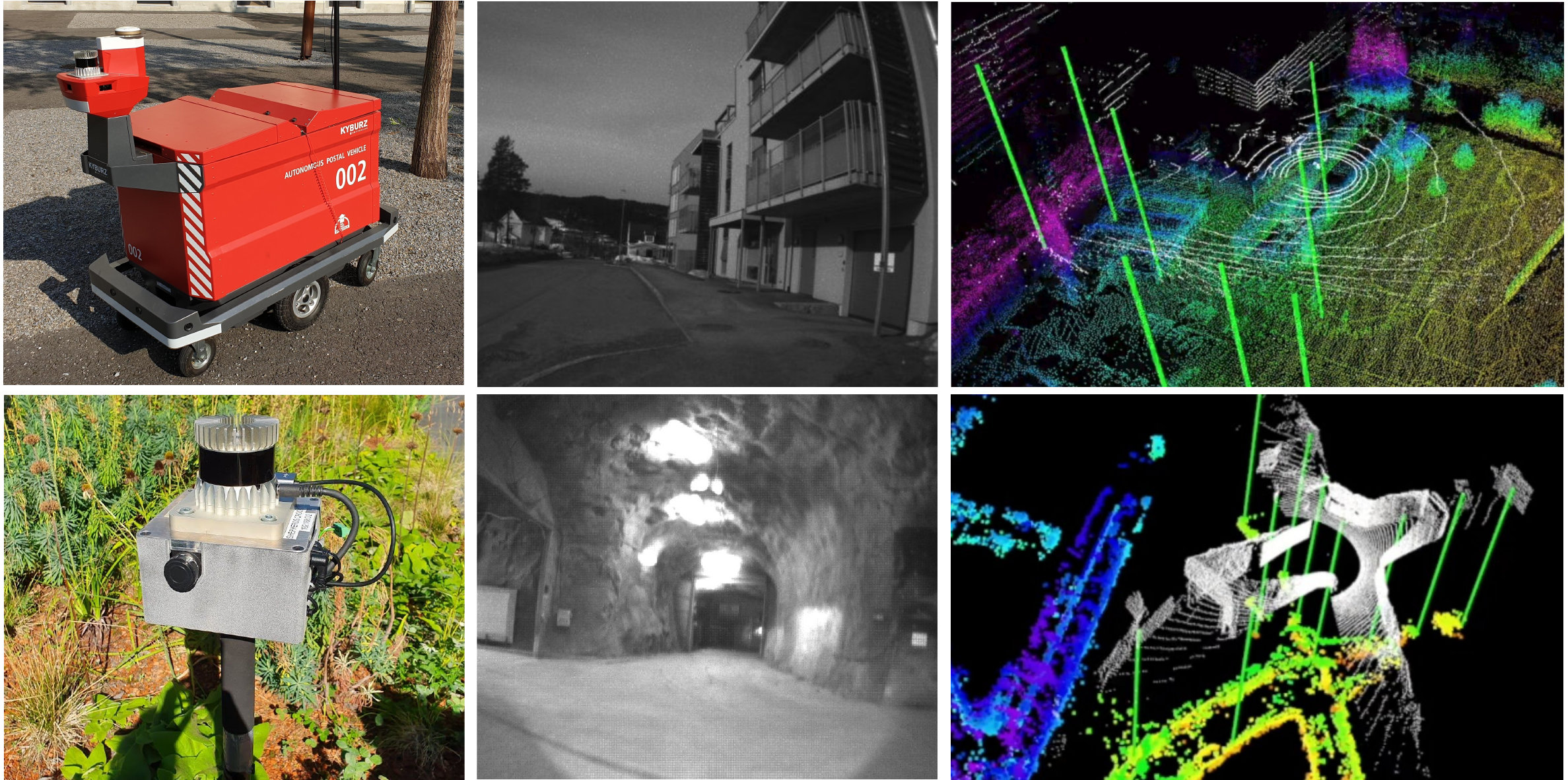}
\caption{Top: traversing a suburban area using a wheeled delivery robot. On the right, a successful localization is depicted. The colored point cloud shows the localization map. The query scan can be seen in white and successfully matched segments are connected with green lines. Bottom: underground mine scenario with a hand-held LiDAR stick. We successfully closed a loop of 5:30 mins length despite the very challenging conditions. We invite the reader to consult our supplementary video for further impressions of the qualitative experiments.}  
\label{fig:qualitative}
\vspace{-4mm}
\end{figure}

In order to test the limits of our approach, we executed qualitative experiments in two real-world environments. The first was performed in an outdoor areal with a wheeled delivery robot. In this 4 minute long map tracking scenario, 92\% of scans passed our geometrical verification step, each counting as a successful localization. Secondly, our approach was tested in a particularly challenging underground mine using a hand-held LiDAR stick.\footnote{The authors thank the Autonomous Systems Lab from ETH Zurich for providing the underground mine dataset.} We were able to successfully close a large loop at an intersection on 191 of 350 scans, despite 15m of drift which had been accumulated at that time. Images showing the experiments are shown in Figure~\ref{fig:qualitative}.

\subsection{LiDAR-Vision Descriptor Fusion}
In order to evaluate the performance of the multimodal descriptor presented in Section~\ref{sec:method_vision} we executed experiments on both the KITTI and NCLT datasets. The multi-modal and LiDAR-only descriptors were trained on the recordings of NCLT of 2012-11-17 and 2012-08-20 and tested on those of 2012-03-25 and 2012-01-15. The test and training data were split according to their position in world frame, as illustrated and detailed in Figure~\ref{fig:data_separation}. For the KITTI dataset, the entire sequence 00 was used as test data, while using the NCLT-trained neural network weights. LiDAR-only and vision-only were chosen as the baselines of the LiDAR-Vision descriptor. For the LiDAR-only case the segment descriptor network presented in Section~\ref{sec:method_lidar} was modified to produce a descriptor of dimension $256 \times 1$ in order to correspond to the size of the multi-modal version. For the vision-only baseline, the segment image patch was input into NetVLAD and the resulting image descriptor was used directly.

\begin{figure}

\centering   
\subfigure {
        \includegraphics[width=0.7\linewidth]{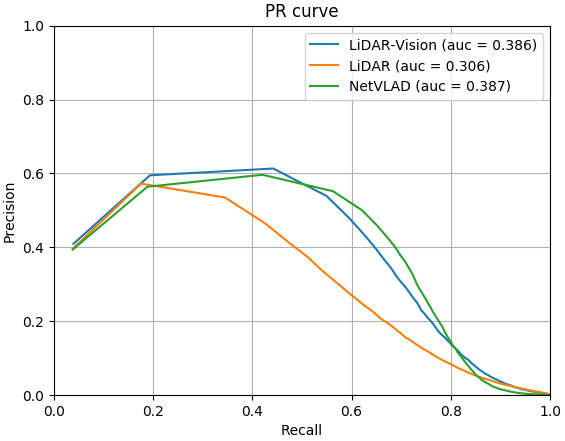}
}
\subfigure {
    \includegraphics[width=0.7\linewidth]{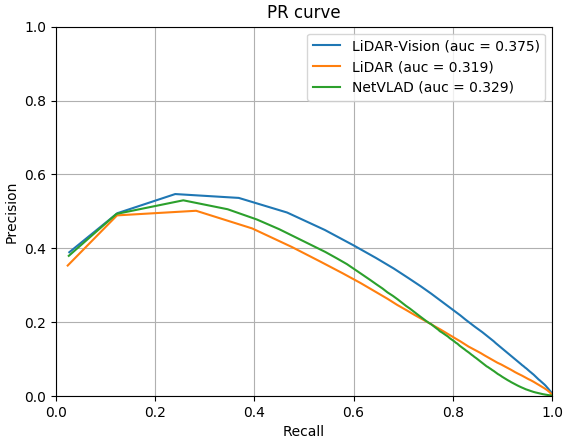}
}

\caption{Precision-recall curves of the descriptor retrieval performance. Top: On KITTI sequence 00, including vision increased the area under the curve by 26\%. Bottom: On the NCLT dataset the performance improved by 17\%.}
\label{fig:pr}
\vspace{-4mm}
\end{figure}

We evaluate our algorithm using a descriptor retrieval scenario, where the Euclidean distance for every pair of descriptors in a dataset is used for classification. Given a threshold $t$ on the descriptor distance, a pair of descriptors is counted as positive if their distance is below this threshold, and as negative otherwise. By comparing this prediction with the ground truth obtained as described in Section~\ref{sec:method_lidar}, the corresponding precision and recall may be computed for different values of the threshold $t$. The resulting precision-recall curves are presented in Figure~\ref{fig:pr}. The experiments show that including vision improves these metrics significantly. We can report increased descriptor retrieval rates of 17-26\% compared to the LiDAR-only baseline. In the case of KITTI, which has almost constant illumination conditions between map and query data, it can be seen that vision strongly boosts the retrieval performance of the multi-modal descriptor. Compared to the vision baseline, the multi-modal descriptor performed worse on KITTI. This may be attributed to the fact that the descriptor network was trained on NCLT, which is particularly challenging for vision with its cross-season setting. This might have led to overfitting, with the network failing to detect when vision would be reliable. For the experiment on the NCLT dataset, the multi-modal descriptor outperforms both the LiDAR and vision baseline.

%% file: chapters/conclusion.tex
\section{Conclusion}
\label{sec:conclusion}
This paper presented \textit{OneShot}, a global localization algorithm based on sparse 3D point cloud segmentation. To the best of our knowledge it is the first segment-based approach able to localize consistently by using only one LiDAR scan at a time. We furthermore introduced a strategy to increase the performance of the segment descriptor by fusing the visual appearance of the underlying structure into the segment descriptor. We report state-of-the-art performance in the robot wake-up scenario on KITTI sequence 00 by localizing after 0.02m on average, compared to 58m by \textit{SegMap}. Finally, it was demonstrated that fusing in the visual appearance of segments resulted in increased descriptor retrieval rates by 17\% to 26\% when compared to using only LiDAR.
